\documentclass[]{seed/bytedance_seed}
\usepackage[T1]{fontenc}

\usepackage[toc,page,header]{appendix}

\usepackage{amsmath,amsfonts,bm}

\def\eqref#1{equation~\ref{#1}}
\def\1{\bm{1}}

\DeclareMathAlphabet{\mathsfit}{\encodingdefault}{\sfdefault}{m}{sl}
\SetMathAlphabet{\mathsfit}{bold}{\encodingdefault}{\sfdefault}{bx}{n}

\usepackage{hyperref}
\usepackage{xurl}
\usepackage{graphicx}
\usepackage{float}
\usepackage{makecell}
\usepackage{url}

\usepackage{minitoc}
\usepackage{amsfonts}
\usepackage{amsmath}
\usepackage{booktabs}
\usepackage{algorithm}
\usepackage{algpseudocode}
\usepackage{cleveref}
\usepackage{pgfplots}
\pgfplotsset{compat=1.18}
\usepackage{xspace}

\newcommand{\CLTK}{Cross-Layer TopK\xspace}
\newcommand{\PLTK}{Per-Layer TopK\xspace}
\newcommand{\CLBTK}{Cross-Layer BatchTopK\xspace}

\title{\CLTK: Dynamically Allocating Sparsity Across Layers in Transcoders}

\author[1,2]{Anyuan Zhuo}
\author[1,\dagger]{Shaohua Zhang}
\author[1,3]{Tianle Li}
\author[1]{Yuan Lin}
\author[1]{Diandian Gu}
\author[1]{\\Jun Wang}
\author[2,\dagger]{Pinyan Lu}

\affiliation[1]{ByteDance Seed}
\affiliation[2]{Shanghai University of Finance and Economics\\}
\affiliation[3]{The University of Hong Kong}
\contribution[\dagger]{Corresponding author}

\abstract{
Cross-Layer Transcoders (CLTs) decompose the activations of large language models into human-interpretable features and trace how these features interact across layers to form circuits, making them a key tool for mechanistic interpretability. However, training CLTs effectively and efficiently is hard. Beyond the usual trade-off between sparsity and reconstruction quality, a CLT has to model all layers at once, which adds a further challenge. The common JumpReLU recipe controls sparsity only indirectly through a penalty, while the direct extension of TopK selects a fixed number of latents in every layer. Yet reconstruction difficulty differs widely across layers, and intermediate layers are much harder than others, so giving every layer the same budget is suboptimal. We propose \CLTK, which sets a single $L_0$ budget for the whole model and dynamically allocates it across layers during training. In a comparison with JumpReLU and other TopK variants on Qwen3-8B-Base, this gives lower FVU while using $20\%$ fewer active latents than JumpReLU. The better reconstruction carries over to circuit tracing, yielding more complete circuits with higher replacement and completeness scores. The fixed per-token budget of \CLTK also makes it easy to pair with sparse decoding, which cuts both computation and communication and gives a $2.9\times$ speedup per training step. We further show that \CLTK follows scaling laws in both latent size and sparsity. We have open‑sourced the weights of our CLT at \url{https://huggingface.co/ByteDance-Seed/Cross-Layer-TopK-CLT}.
}
\date{\today}
\correspondence{\email{{zhangshaohua.cola@bytedance.com}},\email{{lu.pinyan@mail.shufe.edu.cn}}}

\begin{document}
\maketitle

\section{Introduction}

Mechanistic interpretability aims to explain the computation of LLMs in terms of human-interpretable features and the interactions among them, commonly referred to as circuits. A widely used approach is dictionary learning, in which an encoder maps activations to a sparse, over-complete set of features and a decoder reconstructs a target from these features under a sparsity constraint. Sparse autoencoders (SAEs), transcoders, and cross-layer transcoders (CLTs) share this encoder–decoder structure and differ only in their inputs and outputs. SAEs reconstruct an activation from itself and thus extract the features it encodes \citep{gao2025scaling, lieberum2024gemma}. Transcoders instead reconstruct the output of an MLP block from its input, yielding a sparse replacement for the MLP \citep{dunefsky2024transcoders}. CLTs further allow a feature read from the residual stream at layer $l$ to contribute to the MLP outputs of all layers $l' \geq l$, so that computation distributed across many layers can be captured by a single feature. Compared with training one transcoder per layer (per-layer transcoders, PLTs), CLTs have been shown to produce sparser and more interpretable circuits, and have revealed the mechanisms behind real-world model behaviors such as planning, multilingual processing, and jailbreaks \citep{ameisen2025circuit, lindsey2025biology}.

Training dictionaries effectively and efficiently is challenging, since it requires optimizing reconstruction and sparsity at the same time. Sparsity (typically measured by $L_0$, the number of active features) is enforced by the activation function that follows the encoder. Gated and JumpReLU activations control the sparsity level through a penalty term, whereas TopK \citep{gao2025scaling} directly keeps the $k$ largest pre-activations and sets the rest to zero. CLTs add a further difficulty: because a feature at layer $l$ contributes to the MLP outputs of all later layers, all layers must be modeled jointly, and the reconstruction losses of different layers are coupled.

The activation function that enforces this sparsity is therefore a central
design choice. The existing CLT recipe \citep{ameisen2025circuit} uses
JumpReLU, which has practical drawbacks. Its sparsity is obtained indirectly
through a sparsity penalty, and the level reached at the end of training
cannot be predicted from the penalty in advance. Hitting a target sparsity
therefore requires repeated searches over the penalty coefficient, at a
computational cost that is substantial for large CLTs. Even a penalty designed
around a target sparsity \citep{rajamanoharan2024jumping} cannot control it
exactly, and in our experiments JumpReLU also leaves far more latents dead
than TopK (\Cref{sec:reconstruction}).

A natural remedy is to replace the activation function with TopK
\citep{gao2025scaling}, which keeps the $k$ most strongly activated latents
within each layer and thus makes the sparsity exact by construction. This
sacrifices a degree of freedom that JumpReLU enjoys, because layers are not
equally hard to reconstruct. The early layers and the last few layers of an LLM
are typically easier, whereas the middle layers are harder, and activating more
latents at a layer buys a better reconstruction there. Because its penalty is
summed over all layers, JumpReLU is free to redistribute sparsity across layers on its own,
spending more activations where reconstruction is hard. This per-layer
variant, which we call \PLTK, fixes the number of active latents at every layer ahead of time, leaving the model no
such freedom during training.

Motivated by these observations, we propose \CLTK. Rather than selecting the top-$k$ latents in each layer separately, we pool the latents of all $L$ layers and select the $L \cdot k$ with the largest scores for each token. It thus keeps the total sparsity exact, as TopK does, while leaving the allocation across layers to the model, as JumpReLU does. Further analysis shows that \CLTK learns to allocate a larger share of the $L_0$ budget to intermediate layers during training, which leads to its better overall reconstruction. 

We compare \CLTK against two TopK variants in ablations. \PLTK, the direct extension of TopK SAEs to CLTs described above, gives every layer the same budget regardless of how hard it is to reconstruct. \CLBTK, which combines BatchTopK, originally proposed for SAEs \citep{bussmann2024batchtopksparseautoencoders}, with our cross-layer selection, selects the top latents over all tokens in a training batch, which relies on the batch being large and randomly mixed; since no such batch is available at inference, a per-feature activation threshold must be calibrated post hoc on training data. This train–inference mismatch leads to good reconstruction on the training set but a marked drop on the validation set. 

In summary, we propose \CLTK, a simple sparsity mechanism for CLTs, compare it systematically with JumpReLU and other TopK variants, and show that it offers three advantages:
\begin{itemize}
    \item \textbf{Better reconstruction.} By allocating the $L_0$ budget across layers dynamically, \CLTK achieves a lower fraction of variance unexplained (FVU) than all compared methods, even with $20\%$ fewer active latents than JumpReLU, and exhibits scaling laws in both latent size and sparsity.
    \item \textbf{More complete circuits.} The improved reconstruction carries over to circuit tracing, where \CLTK attains higher replacement and completeness scores, indicating better circuit-level interpretability.
    \item \textbf{Simpler sparse decoding.} Because the number of active latents per token is fixed, \CLTK can use sparse decoding directly, without monitoring $L_0$ or handling variable shapes; in our tensor-parallel setup this gives a $2.9\times$ speedup per training step.
\end{itemize}

The rest of the paper is organized as follows. \Cref{sec:related_work} reviews prior work on sparse dictionary learning and circuit analysis. \Cref{sec:architecture} defines \CLTK formally. \Cref{sec:experiments} evaluates it against three other CLT architectures on reconstruction quality (\Cref{sec:reconstruction}), analyzes how it allocates the budget across layers (\Cref{sec:allocation}), evaluates circuit completeness on attribution graphs (\Cref{sec:interpretability}), measures the training speedup that a fixed $L_0$ makes possible (\Cref{sec:efficiency}), and studies scaling laws in latent size and sparsity (\Cref{sec:scaling}). \Cref{sec:conclusion} concludes.

\section{Related Work}
\label{sec:related_work}

\paragraph{Sparse dictionary learning.}
Dictionary learning is motivated by superposition: a layer represents more features than it has neurons \citep{elhage2022superposition}. Sparse autoencoders (SAEs) address this by learning an over-complete dictionary in which each activation is a sparse combination of interpretable directions \citep{bricken2023monosemantic}, and the recipe has since been scaled to frontier models and entire model families \citep{gao2025scaling, lieberum2024gemma}. The mechanism that enforces sparsity has been the main axis of progress. Early SAEs pair a ReLU encoder with an $L_1$ penalty, which shrinks the magnitudes of the latents it selects. Gated SAEs remove that bias by separating the choice of which latents to activate from their magnitude \citep{rajamanoharan2024improving}, and JumpReLU SAEs replace the penalty with a learned per-latent threshold trained through a straight-through estimator \citep{rajamanoharan2024jumping}. TopK instead fixes the number of active latents per token, removing both the penalty coefficient and the threshold \citep{gao2025scaling}. BatchTopK SAEs relax the fixed per-token count into a single budget shared across a batch, so that the quota is allocated dynamically among tokens \citep{bussmann2024batchtopksparseautoencoders}, and Mutual Choice SAEs make a similar batch-level allocation \citep{ayonrinde2024adaptive}. The selection then depends on which inputs share a batch, so a threshold has to be estimated after training before the SAE can be applied to a single input. \CLTK applies the same dynamic allocation along the layer axis, an axis that exists only in cross-layer models. The per-token budget stays fixed, so no post-hoc threshold is needed.

\paragraph{Transcoders and circuit analysis.}
Transcoders learn a sparse map from the input of an MLP to its output, so that the MLP can be replaced rather than merely described \citep{dunefsky2024transcoders}, and their features have been found more interpretable than those of SAEs \citep{paulo2025transcoders}. A natural way to trace circuits is to train one transcoder per layer (per-layer transcoders, PLTs) and chain them. Cross-layer transcoders instead let a feature write to all later layers, yielding a replacement model whose attribution graphs follow directly from the feature activations; \citet{ameisen2025circuit} show that CLTs produce sparser and more interpretable circuits than PLTs, and this has supported circuit-level accounts of frontier models \citep{lindsey2025biology}. Existing CLTs use the JumpReLU architecture~\citep{ameisen2025circuit}, and \citet{lindsey2025landscape} discuss how to train them efficiently at scale, which our sparse decoder builds on. \citet{draye2026cltforge} release an open-source library for training CLTs and computing attribution graphs. We evaluate circuits with the replacement and completeness scores of \citet{ameisen2025circuit}, which measure how much of an attribution graph is carried by features rather than by reconstruction error.

\section{Method}
\label{sec:architecture}

\subsection{Cross-Layer Transcoders}
\label{CLT}

Consider an LLM with $L$ layers. All quantities below are defined for a single token, and we omit the token index. Let $\mathbf{x}^l \in \mathbb{R}^n$ and $\mathbf{y}^l \in \mathbb{R}^n$ denote the input and output of the MLP block at layer $l$, where $n$ is the residual stream dimension. A cross-layer transcoder (CLT)~\citep{ameisen2025circuit} encodes $\mathbf{x}^l$ into a sparse feature activation $\mathbf{a}^l \in \mathbb{R}^m$ at each layer, where $m$ is the latent size, and reconstructs $\mathbf{y}^l$ from the feature activations of layer $l$ and all preceding layers.

\paragraph{Encoder.}
For each layer $l \in \{1, \dots, L\}$, the pre-activation $\mathbf{z}^l \in \mathbb{R}^m$ is
\[
\mathbf{z}^l = \mathbf{W}_{\text{enc}}^l \, \mathbf{x}^l,
\]
where $\mathbf{W}_{\text{enc}}^l \in \mathbb{R}^{m \times n}$ is the encoder weight matrix of layer $l$. Stacking all layers gives $\mathbf{Z} = [\mathbf{z}^1, \dots, \mathbf{z}^L] \in \mathbb{R}^{m \times L}$. The latent activations of all layers, $\mathbf{A} = [\mathbf{a}^1, \dots, \mathbf{a}^L] \in \mathbb{R}^{m \times L}$, are obtained by a sparse activation function $\sigma$:
\[
\mathbf{A} = \sigma(\mathbf{Z}).
\]
We write $\sigma$ as acting on the full matrix $\mathbf{Z}$ because, in general, the activation at one layer may depend on the pre-activations of other layers. For JumpReLU and \PLTK, $\sigma$ simply acts on each column $\mathbf{z}^l$ independently. In all mechanisms we consider, $\sigma$ first rectifies $\mathbf{Z}$ with ReLU, so feature activations are non-negative. The dictionary is over-complete yet each activation is sparse, i.e. $m \gg n$.

\paragraph{Decoder.}
The CLT reconstructs $\mathbf{y}^l$ as
\[
\hat{\mathbf{y}}^l = \sum_{l' \le l} \mathbf{W}_{\text{dec}}^{l' \to l} \, \mathbf{a}^{l'},
\]
where $\mathbf{W}_{\text{dec}}^{l' \to l} \in \mathbb{R}^{n \times m}$ is the decoder weight matrix from the features of layer $l'$ to the output of layer $l$. Following \citet{ameisen2025circuit}, we use no bias terms in either the encoder or the decoder.

\subsection{\CLTK}
\label{sec:cross-layer-topk}
\begin{figure}[t]
    \centering
    \includegraphics[width=0.9\linewidth]{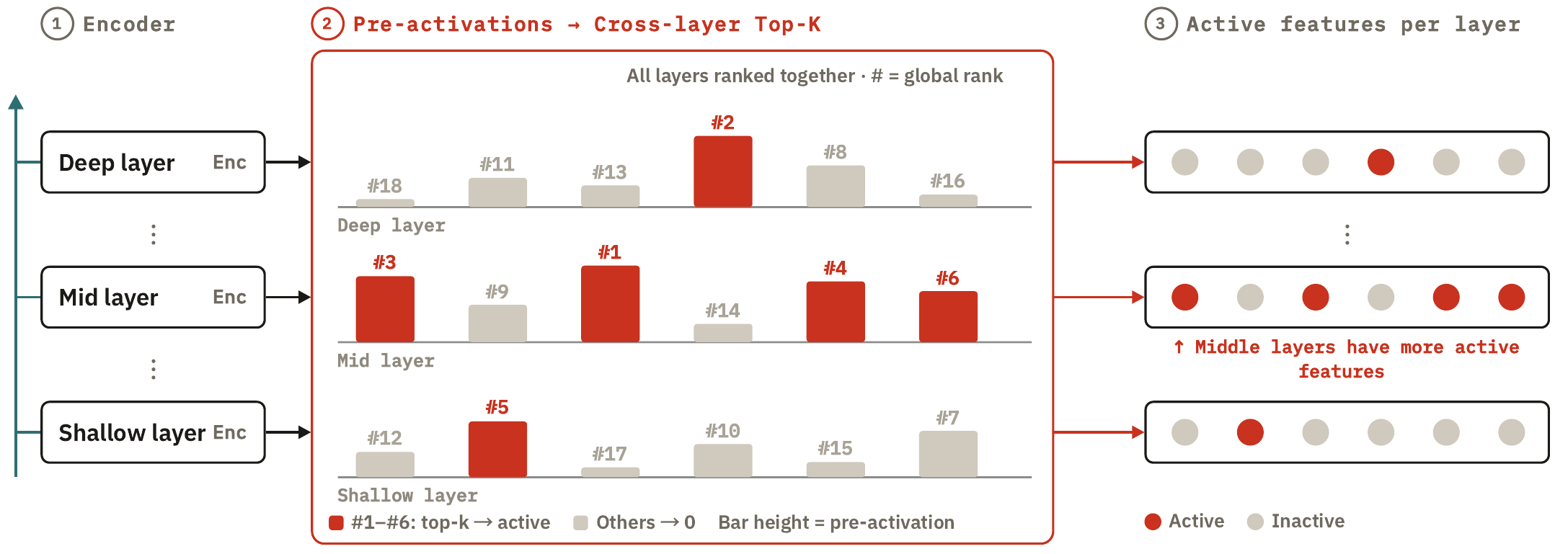}
    \caption{Overview of \CLTK. Each layer's encoder maps the MLP input $\mathbf{x}^l$ to pre-activations $\mathbf{z}^l$. The latents of all $L$ layers are pooled and a single TopK selects the $K = L \cdot k$ with the highest scores, so the number of active latents per layer is decided by the model rather than fixed in advance. Each active latent at layer $l$ then contributes to the reconstruction of layers $l$ through $L$ via its decoder weights.}
    \label{fig:clt_arc}
\end{figure}
\CLTK (\Cref{fig:clt_arc}) is a particular choice of $\sigma$: it ranks all latents of a token across all layers by a scalar score and activates the $K$ largest. The total number of active latents per token is therefore fixed, while their distribution over layers is left to the model. We set $K = L \cdot k$ so that \CLTK has the same total budget as \PLTK with $k$ latents per layer.

Formally, let $s^l_i$ denote the score of latent $i$ at layer $l$, and let $\mathcal{S}$ be the set of the $K$ index pairs $(i, l)$ with the largest scores. \CLTK selects positions by the scores but keeps the rectified values of $\mathbf{Z}$:
\[
a^l_i =
\begin{cases}
\max(z^l_i, 0), & (i, l) \in \mathcal{S}, \\
0, & \text{otherwise,}
\end{cases}
\]

where $a^l_i$ and $z^l_i$ denote the $i$-th entries of $\mathbf{a}^l$ and $\mathbf{z}^l$, indicating the activation and pre-activation of the $i$-th latent in layer $l$, respectively.

Following common practice~\citep{ameisen2025circuit}, we score each latent by its rectified pre-activation scaled by the norm of its decoder weights. Let $\mathbf{w}_i^{l \to l'}$ denote the $i$-th column of $\mathbf{W}_{\text{dec}}^{l \to l'}$. Then
\[
s^l_i = \max(z^l_i, 0) \cdot \Big\lVert \big[\mathbf{w}_i^{l \to l}; \mathbf{w}_i^{l \to l+1}; \dots; \mathbf{w}_i^{l \to L}\big] \Big\rVert_2 ,
\]
which measures the $\ell_2$ norm of the total output that latent $i$ would write to layers $l$ through $L$.

\subsection{Training Objective}
The CLT is trained to reconstruct the MLP output of every layer. The reconstruction loss sums the squared error over all $L$ layers,
\[
\mathcal{L}_{\mathrm{recon}} = \sum_{l=1}^L \lVert \mathbf{y}^l - \hat{\mathbf{y}}^l \rVert_2^2 ,
\]
where the MLP inputs and outputs of each layer are normalized in advance to an expected norm of $\sqrt{n}$, so that the per-layer errors are comparable in scale. To prevent dead latents, we add the auxiliary loss following \citet{gao2025scaling}, adapted to cross-layer selection (\Cref{app:Aux_loss}). The total loss is $\mathcal{L} = \mathcal{L}_{\mathrm{recon}} + \alpha \mathcal{L}_{\mathrm{aux}}$ with coefficient $\alpha$.

\section{Experiments}
\label{sec:experiments}

We evaluate \CLTK on reconstruction quality, circuit completeness, and training efficiency. \Cref{sec:setup} describes the experimental setup. \Cref{sec:reconstruction} compares reconstruction against three other CLT architectures: JumpReLU, \PLTK, and \CLBTK. \Cref{sec:allocation} analyzes how \CLTK allocates the $L_0$ budget across layers. \Cref{sec:interpretability} tests whether better reconstruction leads to more complete circuits, and \Cref{sec:efficiency} measures training efficiency. Finally, \Cref{sec:scaling} shows how reconstruction scales with latent size and sparsity.

\subsection{Experimental Setup}
\label{sec:setup}

We train CLTs on the activations of Qwen3-8B-Base~\citep{yang2025qwen3}. Activations are collected from the Olmo~3 corpus~\citep{olmo2025olmo3}, mixing pre-training text, reasoning chain-of-thought data, and instruction-following SFT data in a $5{:}2{:}1$ ratio. Activations are stored in a buffer and drawn uniformly at random into micro-batches so that each batch mixes tokens from many sequences. Tokens whose activation norm at any layer exceeds $20\times$ the batch mean are discarded as outliers, both in training and in evaluation. Almost all of the discarded tokens are the first token of a sequence.

\subsection{Reconstruction Quality}
\label{sec:reconstruction}

We measure reconstruction by the fraction of variance unexplained (FVU). For layer $l$,
\[
\mathrm{FVU}^l = \frac{\mathbb{E}\big[\lVert \mathbf{y}^l - \hat{\mathbf{y}}^l \rVert_2^2\big]}{\mathbb{E}\big[\lVert \mathbf{y}^l - \bar{\mathbf{y}}^l \rVert_2^2\big]},
\]
where $\bar{\mathbf{y}}^l = \mathbb{E}[\mathbf{y}^l]$. The overall FVU is the average over layers, $\mathrm{FVU} = \frac{1}{L}\sum_{l=1}^L \mathrm{FVU}^l$. All FVUs are measured on a held-out validation set drawn from the same distribution as the training data.

\paragraph{Baselines.}
We compare \CLTK against three other CLT architectures, all with latent size $m = 32{,}768$. The three TopK architectures share identical hyperparameters and activate $K = 360$ latents per token.
\begin{itemize}
    \item \textbf{\PLTK} activates $k$ latents in every layer, with $L \cdot k = K$ so that the total budget matches \CLTK.
    \item \textbf{\CLBTK} extends BatchTopK~\citep{bussmann2024batchtopksparseautoencoders} to CLTs, spending a single budget over all layers and all tokens of a training batch. At inference, a per-latent threshold is estimated from training batches following the original procedure.
    \item \textbf{JumpReLU} is trained with the quadratic $L_0$ penalty of \citet{mcdougall2025gemmascope2}, $\lambda \frac{2}{L_0^{\text{target}}} \big( \lVert \mathbf{a} \rVert_0 - L_0^{\text{target}} \big)^2$, together with the auxiliary loss of \citet{conerly2025dictionarylearning}. Because the penalty vanishes as $L_0$ approaches the target, sparsity stalls above it, ending at an average of $452$ active latents per token.
\end{itemize}
Full hyperparameters are given in \Cref{app:training_setting}.

\paragraph{Results.}
\Cref{tab:reconstruction} summarizes the results. \CLTK attains the lowest FVU of the four architectures. Relative to \PLTK, which differs only in selecting latents within each layer rather than across layers, it lowers FVU from $27.64\%$ to $27.07\%$. This gap is an average over all $36$ layers and understates the gain in the layers that matter most. \CLTK shifts budget from the last layers to the hardest intermediate layers. In layers $15$--$19$, where the FVU of \PLTK exceeds $40\%$, \CLTK lowers it by $1.3$--$1.6$ points (\Cref{fig:layerwise}). \CLBTK reaches $26.98\%$ during training, marginally below \CLTK, but rises to $29.04\%$ at inference once thresholds replace batch-wise selection. JumpReLU reaches only $32.17\%$ despite using more active latents than the other architectures. Since a larger budget works in its favor, this comparison understates the gap. The three TopK architectures leave $3.0$--$3.2\%$ of latents dead, whereas JumpReLU leaves $41.2\%$, so its sparsity budget is concentrated on a much smaller part of the dictionary. 

\begin{table}[t]
\centering
\caption{Reconstruction with $m = 32{,}768$. $L_0$ is the average number of active latents per token. All rows are measured on the validation set, except the \CLBTK training row, which reports the FVU during training with batch-wise selection. Dead latents are those that did not fire in the last $250$ optimizer steps (about $8.2$M tokens).}
\label{tab:reconstruction}
\begin{tabular}{lccc}
\toprule
\textbf{Architecture} & \textbf{$L_0$} & \textbf{FVU} $\downarrow$ & \textbf{Dead latents} $\downarrow$ \\
\midrule
JumpReLU                                          & $452$ & $32.17\%$ & $41.2\%$ \\
\CLBTK (threshold, inference)       & $289$ & $29.04\%$ & --- \\
\PLTK                                     & $360$ & $27.64\%$ & $3.0\%$ \\
\CLTK (ours)                            & $360$ & $\mathbf{27.07\%}$ & $3.2\%$ \\
\CLBTK (batch selection, training)  & $360$ & $\underline{26.98\%}$ & $3.0\%$ \\
\bottomrule
\end{tabular}
\end{table}

\paragraph{\CLBTK degrades at inference.}
The threshold of each latent is estimated by averaging, over training batches, the smallest activation of that latent that survived selection. Batches in which the latent never fires contribute nothing, and latent activations are bursty, so the estimate is taken from exactly the batches in which the latent was most active. The threshold is therefore biased upwards, most severely for rare latents, and \CLBTK activates fewer latents at inference than in training.

\paragraph{JumpReLU cannot reach the target sparsity.}
The quadratic penalty decays to zero as sparsity approaches the target, so near the target it is dominated by the reconstruction loss and sparsity stalls above the target. We also find JumpReLU sensitive to the penalty coefficient $\lambda$: if sparsity decreases too fast, a large fraction of latents die, so $\lambda$ has to be tuned so that sparsity approaches the target gradually, which requires repeated training runs.

\subsection{Cross-Layer Allocation}
\label{sec:allocation}

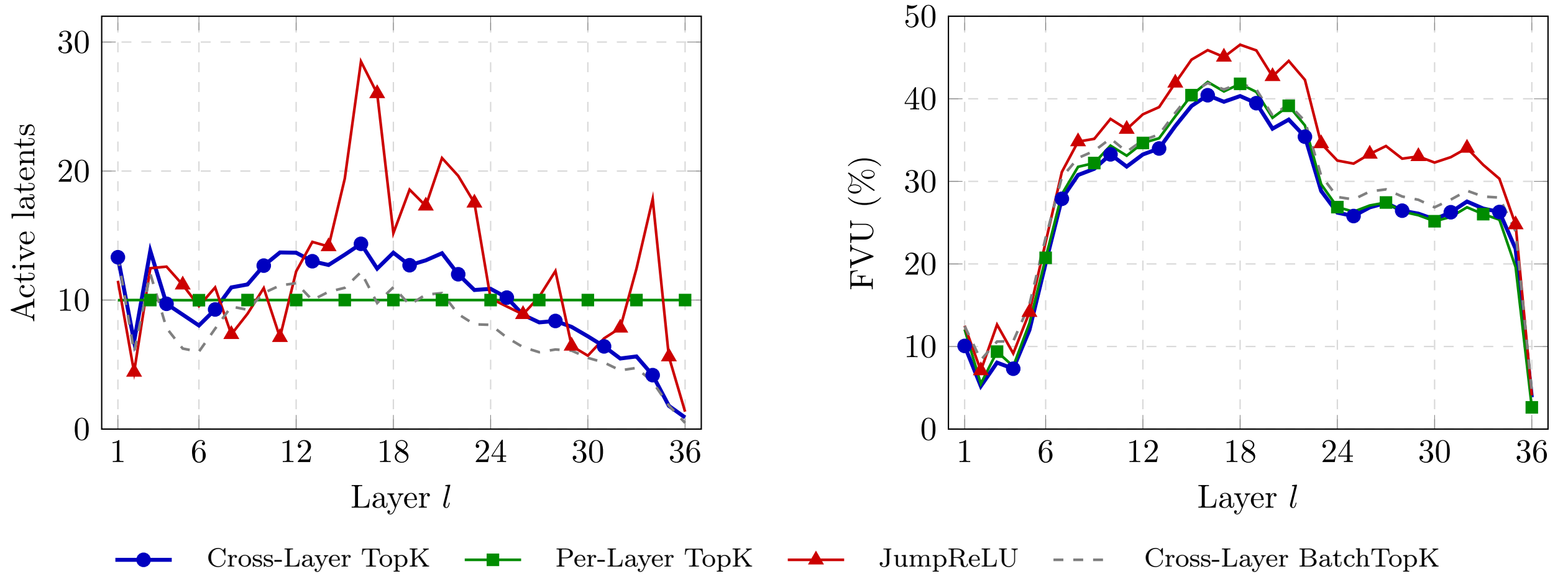
\begin{figure}[H]
    \centering
    \begin{tikzpicture}
        \begin{axis}[
            width=0.48\linewidth, height=0.36\linewidth,
            xmin=0, xmax=37, ymin=0, ymax=32,
            xtick={1,6,12,18,24,30,36},
            xlabel={Layer $l$}, ylabel={Active latents},
            grid=major, grid style={dashed, gray!30},
            legend to name=layerwise-legend, legend columns=4,
            legend style={font=\footnotesize, draw=none, column sep=8pt},
        ]
        \addplot[very thick, blue!75!black, mark=*, mark size=1.6pt, mark repeat=3] coordinates {
            (1,13.32) (2,6.79) (3,13.81) (4,9.71) (5,8.87) (6,8.03)
            (7,9.27) (8,10.98) (9,11.21) (10,12.67) (11,13.69) (12,13.67)
            (13,13.01) (14,12.72) (15,13.52) (16,14.36) (17,12.44) (18,13.67)
            (19,12.70) (20,13.08) (21,13.63) (22,12.00) (23,10.78) (24,10.87)
            (25,10.19) (26,8.86) (27,8.27) (28,8.38) (29,7.92) (30,7.19)
            (31,6.41) (32,5.47) (33,5.63) (34,4.18) (35,1.80) (36,0.90)
        };
        \addlegendentry{\CLTK}
        \addplot[thick, green!55!black, mark=square*, mark size=1.5pt, mark repeat=3, mark phase=3] coordinates {
            (1,9.99) (2,10.00) (3,10.00) (4,10.00) (5,9.99) (6,9.99)
            (7,10.00) (8,10.00) (9,10.00) (10,10.00) (11,10.00) (12,10.00)
            (13,10.00) (14,10.00) (15,10.00) (16,10.00) (17,10.00) (18,10.00)
            (19,10.00) (20,10.00) (21,10.00) (22,10.00) (23,10.00) (24,10.00)
            (25,10.00) (26,10.00) (27,10.00) (28,10.00) (29,10.00) (30,10.00)
            (31,10.00) (32,10.00) (33,10.00) (34,10.00) (35,10.00) (36,10.00)
        };
        \addlegendentry{\PLTK}
        \addplot[thick, red!80!black, mark=triangle*, mark size=2pt, mark repeat=3, mark phase=2] coordinates {
            (1,11.49) (2,4.42) (3,12.47) (4,12.59) (5,11.18) (6,9.53)
            (7,10.98) (8,7.34) (9,8.94) (10,10.93) (11,7.11) (12,12.22)
            (13,14.51) (14,14.15) (15,19.41) (16,28.49) (17,26.01) (18,15.19)
            (19,18.56) (20,17.30) (21,21.01) (22,19.63) (23,17.53) (24,10.12)
            (25,9.50) (26,8.89) (27,10.26) (28,12.25) (29,6.44) (30,5.69)
            (31,7.03) (32,7.84) (33,12.43) (34,17.83) (35,5.61) (36,1.35)
        };
        \addlegendentry{JumpReLU}
        \addplot[thick, dashed, black!50, mark=none] coordinates {
            (1,13.40) (2,5.96) (3,12.05) (4,7.83) (5,6.24) (6,6.00)
            (7,7.78) (8,9.50) (9,9.25) (10,10.55) (11,11.13) (12,11.31)
            (13,10.00) (14,10.64) (15,10.95) (16,12.17) (17,9.77) (18,11.00)
            (19,9.64) (20,10.42) (21,10.54) (22,8.91) (23,8.12) (24,8.08)
            (25,7.10) (26,6.35) (27,5.96) (28,6.17) (29,6.08) (30,5.52)
            (31,5.16) (32,4.55) (33,4.74) (34,3.66) (35,1.82) (36,0.48)
        };
        \addlegendentry{\CLBTK}
        \end{axis}
    \end{tikzpicture}
    \hfill
    \begin{tikzpicture}
        \begin{axis}[
            width=0.48\linewidth, height=0.36\linewidth,
            xmin=0, xmax=37, ymin=0, ymax=50,
            xtick={1,6,12,18,24,30,36}, ytick={0,10,20,30,40,50},
            xlabel={Layer $l$}, ylabel={FVU (\%)},
            grid=major, grid style={dashed, gray!30},
        ]
        \addplot[very thick, blue!75!black, mark=*, mark size=1.6pt, mark repeat=3] coordinates {
            (1,10.07) (2,5.20) (3,8.05) (4,7.31) (5,12.00) (6,20.09)
            (7,27.89) (8,30.77) (9,31.52) (10,33.27) (11,31.79) (12,33.26)
            (13,33.98) (14,36.70) (15,39.13) (16,40.43) (17,39.64) (18,40.33)
            (19,39.47) (20,36.38) (21,37.48) (22,35.41) (23,28.84) (24,26.21)
            (25,25.80) (26,26.85) (27,27.39) (28,26.45) (29,26.09) (30,25.40)
            (31,26.26) (32,27.56) (33,26.71) (34,26.29) (35,21.83) (36,3.85)
        };
        \addplot[thick, green!55!black, mark=square*, mark size=1.5pt, mark repeat=3, mark phase=3] coordinates {
            (1,12.07) (2,5.49) (3,9.38) (4,7.60) (5,12.82) (6,20.74)
            (7,28.45) (8,31.76) (9,32.22) (10,34.33) (11,33.11) (12,34.65)
            (13,35.22) (14,37.88) (15,40.44) (16,42.04) (17,40.89) (18,41.81)
            (19,40.83) (20,37.70) (21,39.15) (22,36.77) (23,29.65) (24,26.88)
            (25,26.33) (26,27.08) (27,27.44) (28,26.34) (29,25.90) (30,25.17)
            (31,25.69) (32,26.86) (33,26.03) (34,25.36) (35,19.62) (36,2.63)
        };
        \addplot[thick, red!80!black, mark=triangle*, mark size=2pt, mark repeat=3, mark phase=2] coordinates {
            (1,12.51) (2,7.08) (3,12.65) (4,9.16) (5,14.15) (6,22.65)
            (7,31.14) (8,34.81) (9,35.15) (10,37.56) (11,36.32) (12,38.12)
            (13,38.99) (14,41.92) (15,44.74) (16,45.89) (17,45.06) (18,46.55)
            (19,45.86) (20,42.71) (21,44.59) (22,42.29) (23,34.58) (24,32.52)
            (25,32.15) (26,33.31) (27,34.28) (28,32.75) (29,33.02) (30,32.28)
            (31,32.93) (32,34.03) (33,32.00) (34,30.32) (35,24.76) (36,4.09)
        };
        \addplot[thick, dashed, black!50, mark=none] coordinates {
            (1,12.47) (2,8.37) (3,10.60) (4,10.65) (5,15.12) (6,23.19)
            (7,30.49) (8,32.83) (9,33.69) (10,35.19) (11,33.59) (12,34.98)
            (13,35.68) (14,38.31) (15,40.75) (16,41.90) (17,41.13) (18,41.89)
            (19,41.14) (20,37.98) (21,39.34) (22,37.30) (23,30.70) (24,28.11)
            (25,27.81) (26,28.77) (27,29.03) (28,28.17) (29,27.77) (30,26.86)
            (31,27.79) (32,28.87) (33,28.16) (34,28.04) (35,23.94) (36,4.67)
        };
        \end{axis}
    \end{tikzpicture}\\[2pt]
    \pgfplotslegendfromname{layerwise-legend}
    \caption{Per-layer allocation (left) and FVU (right). \PLTK is flat by construction; the architectures that can redistribute concentrate their budget on the intermediate layers, where FVU is highest, and withdraw it from the last layers.}
    \label{fig:layerwise}
\end{figure}

The comparison above shows that freeing the allocation of latents across layers lowers the FVU. We now examine how \CLTK uses this freedom. \Cref{fig:layerwise} reports, for every layer, the average number of active latents it receives (left) and the FVU it attains (right).

\paragraph{Intermediate layers are harder to reconstruct.}
Under \PLTK, which gives every layer the same $10$ latents, the FVU rises from about $5$--$12\%$ in the first few layers to over $40\%$ in layers $15$--$19$, then falls back to under $4\%$ in the last layer. Reconstruction difficulty is thus far from uniform across depth, and a uniform budget is poorly matched to it.

\paragraph{\CLTK moves the budget toward the hard layers.}
With the same total of $360$ latents per token, \CLTK allocates $12$--$14$ latents to most of layers $10$--$22$, peaking at $14.36$ in layer $16$, and progressively withdraws from the last layers, down to $0.90$ in layer $36$. Over the intermediate layers, the budget follows the FVU profile of the fixed-budget baseline and concentrates where reconstruction is hardest: across layers $5$--$27$, the number of latents a layer receives under \CLTK correlates with its FVU under \PLTK at $r = 0.85$. The first layers deviate from this pattern and receive more latents than their low FVU would suggest, e.g.\ $13.8$ in layer $3$, consistent with the observation of \citet{lange2026clt} that early-layer features of a CLT can write to all later layers at little extra cost. For the same reason, the outputs of the last layers are partly reconstructed by latents from earlier layers, so these layers receive few latents of their own even though their FVU under \PLTK remains around $25\%$. JumpReLU, which is also free to redistribute, arrives at a similar profile with a much larger swing (from $1.35$ to $28.49$ latents) and peaks in the same layer. \CLBTK follows the same profile but shifted downward, reflecting its reduced budget of about $289$ latents at inference.

\paragraph{The reallocation is a trade.}
\CLTK reconstructs $27$ of the $36$ layers better than \PLTK, with the largest gain of $2.0$ points in layer $1$. The nine exceptions are exactly the last nine layers, from which it withdrew most of its budget, with the largest loss of $2.2$ points in layer $35$. Averaged over layers the trade is favorable, as the overall FVU in \Cref{tab:reconstruction} shows. JumpReLU is worse than \CLTK in every layer despite its larger budget.

\subsection{Circuit Completeness}
\label{sec:interpretability}
CLTs are built for circuit tracing~\citep{ameisen2025circuit}. Circuit tracing substitutes the CLT for the MLP blocks of the model and expresses the computation behind a prediction as an \emph{attribution graph}, whose nodes are the token embeddings, the active CLT features, and the target logit, and whose edges carry the direct linear effect of one node on another. Since the CLT does not reconstruct the MLP outputs exactly, an error node is added at every layer and token position to absorb the residual. Paths through feature nodes can be inspected feature by feature, whereas error nodes are opaque. The fraction of influence routed through features therefore measures how much of the computation the attribution graph explains in terms of features, and \citet{ameisen2025circuit} quantify it at two granularities. The \emph{completeness score} is local: for each node, the fraction of its incoming influence that comes from non-error nodes, averaged over nodes weighted by their influence on the target logit. The \emph{replacement score} is global: the fraction of path influence from the token embeddings to the target logit that passes only through non-error nodes. Both range from $0$ to $1$, higher being better; precise definitions are given in \Cref{app:graph_scores}.

\begin{table}[H]
\centering
\caption{Replacement and completeness scores, averaged over the same $100$ prompts for all architectures. $L_0$ is the average number of active latents per token at inference. Higher is better.}
\label{tab:graph_scores}
\begin{tabular}{lccc}
\toprule
\textbf{Architecture} & \textbf{$L_0$} & \textbf{Replacement} $\uparrow$ & \textbf{Completeness} $\uparrow$ \\
\midrule
JumpReLU              & $452$ & $0.589$ & $0.897$ \\
\PLTK         & $360$ & $0.577$ & $0.894$ \\
\CLBTK  & $289$ & $0.578$ & $0.889$ \\
\CLTK (ours) & $360$ & $\mathbf{0.617}$ & $\mathbf{0.902}$ \\
\bottomrule
\end{tabular}
\end{table}

We sample $100$ prompts from a held-out validation set drawn from the same distribution as the training data. Following \citet{ameisen2025circuit}, we exclude prompts whose next token is trivial to predict, keeping a prompt only if Qwen3-8B-Base predicts the ground-truth token under greedy decoding, the token's unigram frequency in the pre-training corpus is below $10^{-4}$, the token does not appear earlier in the prompt, and the cross-entropy of the prediction exceeds $0.2$. The same $100$ prompts are used for all four architectures from \Cref{sec:reconstruction} ($m = 32{,}768$). Building and scoring an attribution graph requires a full forward pass of the replacement model followed by attribution over every active feature at every token position, so $100$ prompts is a substantial evaluation.

\Cref{tab:graph_scores} reports the scores averaged over the $100$ prompts. \CLTK attains the highest replacement and completeness scores of the four architectures. At the same budget of $360$ latents per token, it exceeds \PLTK by $0.04$ on replacement score and $0.008$ on completeness score. \CLBTK, whose post-hoc thresholds leave it with only $289$ active latents at inference, scores about the same as \PLTK on replacement and lowest on completeness. JumpReLU ranks second on replacement score despite the worst FVU; this is explained by its larger budget of $452$ active latents per token, which gives the graph more feature paths to route influence through. That \CLTK still exceeds it on both scores with $20\%$ fewer active latents indicates that the gain comes from better reconstruction rather than from a larger budget.

\subsection{Training Efficiency}
\label{sec:efficiency}
\CLTK activates exactly $K$ latents per token, so the sparsity pattern of the decoder input is known before decoding and has a fixed size. This makes it simple to pair with a sparse decoder: the $K$ active latents can be exchanged across ranks and decoded directly, with no dynamic shapes or padding. JumpReLU offers no such guarantee, since its number of active latents varies per token, and \CLBTK fixes it only at the batch level.

\begin{figure}[t]
    \centering
    \includegraphics[width=0.95\linewidth]{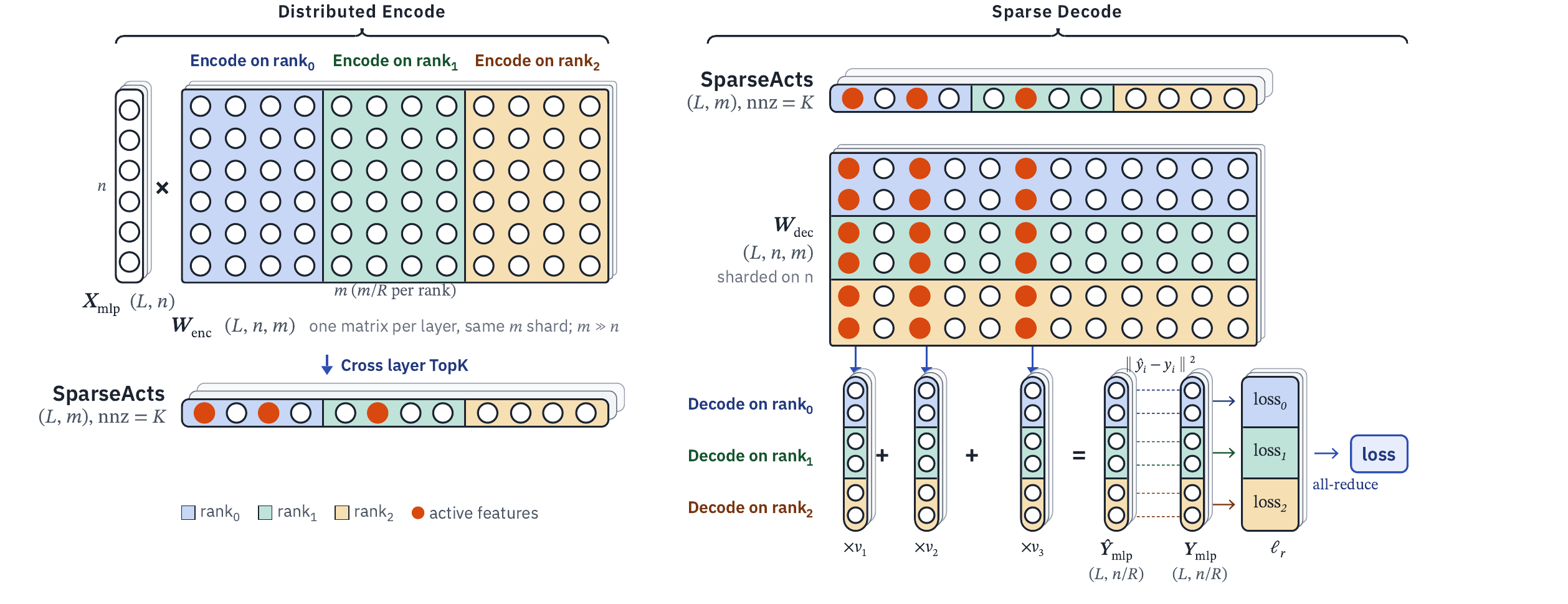}
    \caption{Sparse decoding. The encoder is sharded along the latent axis, the decoder along the model dimension, and only the $K$ active latents of each token are exchanged across ranks, avoiding the dense reconstruction all-reduce. The batch dimension is omitted for clarity.}
    \label{fig:sparse_decode}
\end{figure}

We pair \CLTK with sparse decoding for tensor-parallel training (\Cref{fig:sparse_decode}), following the design notes of \citet{lindsey2025landscape}. The encoder is sharded along the latent axis, the decoder along the model dimension, and only the $K$ active latents of each token are exchanged across the $R$ tensor-parallel ranks. This removes the dense all-reduce of the reconstruction that dominates communication when both encoder and decoder are sharded along the latent axis. Sparse decoding has also been used for JumpReLU CLTs, where training monitors $L_0$ and switches from dense to sparse decoding once it falls below a crossover point~\citep{draye2026cltforge}. With \CLTK the number of active latents per token is set by construction, so no monitoring or switching is needed. We compare three numerically equivalent kernels for building the reconstruction from the active latents:
\begin{itemize}
    \item \texttt{dense} scatters the active latents back into a dense buffer and runs the original dense matmul. It is not sparsity-aware, but compared with sharding both encoder and decoder along the latent axis it reads only $1/R$ of the decoder weights per rank and avoids the dense all-reduce.
    \item \texttt{gather} gathers the selected rows of the decoder and accumulates them with \texttt{index\_add}. It reads only the active rows, but materializes a temporary of size proportional to the number of active latents and relies on atomics.
    \item \texttt{embedding\_bag} fuses the gather, the scaling, and the per-token sum into one kernel. It materializes no temporary and needs no atomics.
\end{itemize}
\Cref{app:sparse_decode} gives the algorithm.

\Cref{tab:sparse_decode_time} compares the three kernels against the baseline that shards both encoder and decoder along the latent axis, on $128$ GPUs with $B = 8192$ and $k = 10$. The \texttt{dense} kernel alone nearly halves the step time through reduced memory traffic and communication; the sparsity-aware kernels reduce it further, and \texttt{embedding\_bag} is fastest, cutting compute time by $65\%$ and communication time by $67\%$ for a $2.9\times$ speedup per training step.

\begin{table}[t]
\centering
\caption{Step time on $128$ GPUs at $k = 10$, $B = 8192$. Step time is split into compute (including memory access) and communication.}
\label{tab:sparse_decode_time}
\begin{tabular}{lcccc}
\toprule
\textbf{Scheme} & \textbf{Step (s)} & \textbf{Compute (s)} & \textbf{Comm.\ (s)} & \textbf{Speedup} \\
\midrule
Baseline (latent-sharded)              & $1.025$ & $0.800$ & $0.224$ & $1.0\times$ \\
Sparse decode, \texttt{dense}          & $0.552$ & $0.451$ & $0.101$ & $1.9\times$ \\
Sparse decode, \texttt{gather}         & $0.436$ & $0.334$ & $0.102$ & $2.4\times$ \\
Sparse decode, \texttt{embedding\_bag} & $\mathbf{0.355}$ & $\mathbf{0.281}$ & $\mathbf{0.074}$ & $\mathbf{2.9\times}$ \\
\bottomrule
\end{tabular}
\end{table}

\subsection{Scaling Laws}
\label{sec:scaling}
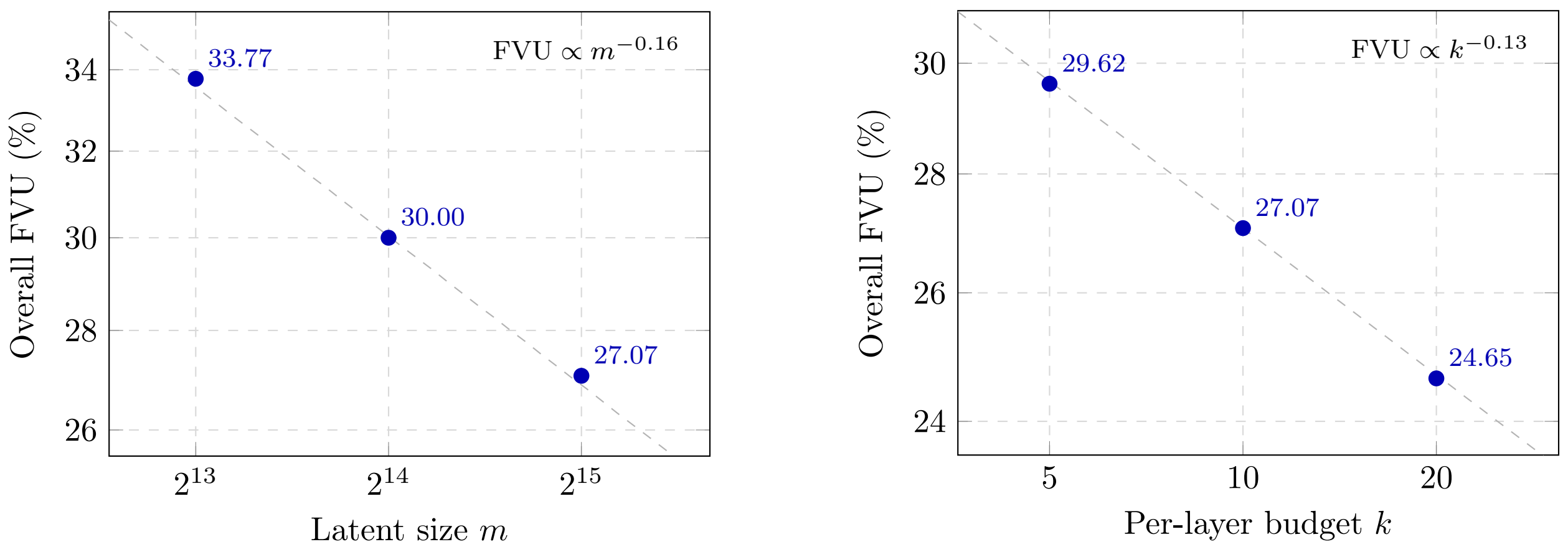
\begin{figure}[H]
    \centering
    \begin{tikzpicture}
        \begin{axis}[
            width=0.48\linewidth, height=0.38\linewidth,
            xmode=log, log basis x={2}, ymode=log, log basis y={10},
            log ticks with fixed point,
            xmin=6000, xmax=52000, ymin=25.5, ymax=35.5,
            xtick={8192,16384,32768}, xticklabels={$2^{13}$, $2^{14}$, $2^{15}$},
            ytick={26,28,30,32,34}, yminorticks=false,
            xlabel={Latent size $m$}, ylabel={Overall FVU (\%)},
            grid=major, grid style={dashed, gray!30},
            every node near coord/.append style={anchor=south west, font=\footnotesize},
        ]
        \addplot[dashed, gray!60, domain=6000:52000, samples=40] {141.9*x^(-0.160)};
        \addplot[thick, blue!70!black, mark=*, mark size=2pt, only marks, nodes near coords, point meta=explicit symbolic]
            coordinates { (8192,33.77) [$33.77$] (16384,30.00) [$30.00$] (32768,27.07) [$27.07$] };
        \node[anchor=north east, font=\footnotesize] at (rel axis cs:0.97,0.97) {$\mathrm{FVU} \propto m^{-0.16}$};
        \end{axis}
    \end{tikzpicture}
    \hfill
    \begin{tikzpicture}
        \begin{axis}[
            width=0.48\linewidth, height=0.38\linewidth,
            xmode=log, log basis x={2}, ymode=log, log basis y={10},
            log ticks with fixed point,
            xmin=3.6, xmax=31, ymin=23.5, ymax=31,
            xtick={5,10,20}, xticklabels={$5$, $10$, $20$},
            ytick={24,26,28,30}, yminorticks=false,
            xlabel={Per-layer budget $k$}, ylabel={Overall FVU (\%)},
            grid=major, grid style={dashed, gray!30},
            every node near coord/.append style={anchor=south west, font=\footnotesize},
        ]
        \addplot[dashed, gray!60, domain=3.6:31, samples=40] {36.68*x^(-0.132)};
        \addplot[thick, blue!70!black, mark=*, mark size=2pt, only marks, nodes near coords, point meta=explicit symbolic]
            coordinates { (5,29.62) [$29.62$] (10,27.07) [$27.07$] (20,24.65) [$24.65$] };
        \node[anchor=north east, font=\footnotesize] at (rel axis cs:0.97,0.97) {$\mathrm{FVU} \propto k^{-0.13}$};
        \end{axis}
    \end{tikzpicture}
    \caption{Scaling of \CLTK. Left: overall FVU against latent size $m$ at $k = 10$. Right: overall FVU against the per-layer budget $k$ at $m = 2^{15}$, where each model activates $K = L \cdot k$ latents per token. Both axes are logarithmic, so a power law appears as a straight line; dashed lines are least-squares fits in log--log space.}
    \label{fig:scaling}
\end{figure}

We study how \CLTK scales along each of its two axes. \Cref{fig:scaling} (left) varies the latent size $m$ from $2^{13}$ to $2^{15}$ at a fixed per-layer budget $k = 10$; \Cref{fig:scaling} (right) varies $k$ from $5$ to $20$ at a fixed $m = 2^{15}$. All other hyperparameters are held fixed and every point is a separately trained model. In both cases the overall FVU follows a power law, $\mathrm{FVU} \propto m^{-0.16}$ and $\mathrm{FVU} \propto k^{-0.13}$, so doubling the latent size or the budget lowers FVU by about $11\%$ and $9\%$ respectively. These exponents are of the same order as those implied by the scaling law that \citet{gao2025scaling} fit for TopK SAEs over a comparable range. \citet{ameisen2025circuit} report Pareto improvements as CLT dictionaries grow, but because JumpReLU controls sparsity only through a penalty, $L_0$ shifts along with latent size and the two effects cannot be separated; \CLTK fixes $L_0$ exactly, which lets us sweep each axis in isolation.

\section{Conclusion}
\label{sec:conclusion}

We presented \CLTK, a sparsity mechanism for cross-layer transcoders that selects the top-$K$ latents over all layers of a token at once. The number of active latents per token is fixed by construction, while their allocation across layers is left to the model.  The selection depends on the token alone, so no threshold has to be estimated after training. On Qwen3-8B-Base, \CLTK learns to concentrate its budget on the intermediate layers that are hardest to reconstruct, attains lower FVU than \PLTK at the same sparsity and than JumpReLU at a looser one, and yields more complete circuits, with higher replacement and completeness scores. Its fixed budget also makes sparse decoding straightforward to apply, cutting training step time by $2.9\times$ in our setup, and its reconstruction follows scaling laws in both latent size and sparsity.

Extending TopK from SAEs to CLTs is an intuitive step, and variants of it have appeared as implementation options in prior work and open-source libraries, but they have not been studied in depth or compared systematically. Our results show that the choice among them matters, and that selecting across layers is what makes TopK work well for CLTs. A CLT reconstructs every layer at once, and its decoder grows with the square of the number of layers, so each training run of a large CLT consumes substantial compute. This also kept us from scaling to larger size. More broadly, we hope this work helps make circuit tracing practical for frontier LLMs, advancing our understanding of how they compute and supporting the development of safer AI systems.

%不需要目录就注释掉 注意目录不要和第一页放在一块 要有\newpage
%\newpage
%\tableofcontents
%\newpage

\clearpage

\bibliography{iclr2027_conference}

@article{yang2025qwen3,
  title={Qwen3 technical report},
  author={Yang, An and Li, Anfeng and Yang, Baosong and Zhang, Beichen and Hui, Binyuan and Zheng, Bo and Yu, Bowen and Gao, Chang and Huang, Chengen and Lv, Chenxu and others},
  journal={arXiv preprint arXiv:2505.09388},
  year={2025}
}

@misc{olmo2025olmo3,
title={Olmo 3},
author={Team Olmo and Allyson Ettinger and Amanda Bertsch and Bailey Kuehl and David Graham and David Heineman and Dirk Groeneveld and Faeze Brahman and Finbarr Timbers and Hamish Ivison and Jacob Morrison and Jake Poznanski and Kyle Lo and Luca Soldaini and Matt Jordan and Mayee Chen and Michael Noukhovitch and Nathan Lambert and Pete Walsh and Pradeep Dasigi and Robert Berry and Saumya Malik and Saurabh Shah and Scott Geng and Shane Arora and Shashank Gupta and Taira Anderson and Teng Xiao and Tyler Murray and Tyler Romero and Victoria Graf and Akari Asai and Akshita Bhagia and Alexander Wettig and Alisa Liu and Aman Rangapur and Chloe Anastasiades and Costa Huang and Dustin Schwenk and Harsh Trivedi and Ian Magnusson and Jaron Lochner and Jiacheng Liu and Lester James V. Miranda and Maarten Sap and Malia Morgan and Michael Schmitz and Michal Guerquin and Michael Wilson and Regan Huff and Ronan Le Bras and Rui Xin and Rulin Shao and Sam Skjonsberg and Shannon Zejiang Shen and Shuyue Stella Li and Tucker Wilde and Valentina Pyatkin and Will Merrill and Yapei Chang and Yuling Gu and Zhiyuan Zeng and Ashish Sabharwal and Luke Zettlemoyer and Pang Wei Koh and Ali Farhadi and Noah A. Smith and Hannaneh Hajishirzi},
year={2025},
eprint={2512.13961},
archivePrefix={arXiv},
primaryClass={cs.CL},
url={https://arxiv.org/abs/2512.13961},
}

@inproceedings{gao2025scaling,
  title={Scaling and evaluating sparse autoencoders},
  author={Gao, Leo and Dupre la Tour, Tom and Tillman, Henk and Goh, Gabriel and Troll, Rajan and Radford, Alec and Sutskever, Ilya and Leike, Jan and Wu, Jeffrey},
  booktitle={International Conference on Learning Representations},
  volume={2025},
  pages={26721--26754},
  year={2025}
}

@misc{loshchilov2019decoupledweightdecayregularization,
      title={Decoupled Weight Decay Regularization}, 
      author={Ilya Loshchilov and Frank Hutter},
      year={2019},
      eprint={1711.05101},
      archivePrefix={arXiv},
      primaryClass={cs.LG},
      url={https://arxiv.org/abs/1711.05101}, 
}

@misc{conerly2025dictionarylearning,
  author       = {Conerly, Tom and Cunningham, Hoagy and Templeton, Adly and
                  Lindsey, Jack and Hosmer, Basil and Jermyn, Adam},
  title        = {Dictionary Learning Optimization Techniques},
  howpublished = {Transformer Circuits Thread, Circuits Updates --- January 2025},
  year         = {2025},
  month        = jan,
  url          = {https://transformer-circuits.pub/2025/january-update/index.html#DL},
  note         = {Accessed: 2026-07-08}
}

@misc{ameisen2025circuit,
  title        = {Circuit Tracing: Revealing Computational Graphs in Language Models},
  author       = {Ameisen, Emmanuel and Lindsey, Jack and Pearce, Adam and
                  Gurnee, Wes and Turner, Nicholas L. and Chen, Brian and
                  Citro, Craig and Abrahams, David and Carter, Shan and
                  Hosmer, Basil and Marcus, Jonathan and Sklar, Michael and
                  Templeton, Adly and Bricken, Trenton and McDougall, Callum and
                  Cunningham, Hoagy and Henighan, Tom and Jermyn, Adam and
                  Jones, Andy and Persic, Andrew and Qi, Zhenyi and
                  Thompson, T. Ben and Zimmerman, Sam and Rivoire, Kelley and
                  Conerly, Thomas and Olah, Chris and Batson, Joshua},
  year         = {2025},
  url          = {https://transformer-circuits.pub/2025/attribution-graphs/methods.html},
  note         = {Transformer Circuits Thread}
}

@misc{bussmann2024batchtopksparseautoencoders,
      title={BatchTopK Sparse Autoencoders}, 
      author={Bart Bussmann and Patrick Leask and Neel Nanda},
      year={2024},
      eprint={2412.06410},
      archivePrefix={arXiv},
      primaryClass={cs.LG},
      url={https://arxiv.org/abs/2412.06410}, 
}

@inproceedings{lieberum2024gemma,
  title={Gemma scope: Open sparse autoencoders everywhere all at once on gemma 2},
  author={Lieberum, Tom and Rajamanoharan, Senthooran and Conmy, Arthur and Smith, Lewis and Sonnerat, Nicolas and Varma, Vikrant and Kram{\'a}r, J{\'a}nos and Dragan, Anca and Shah, Rohin and Nanda, Neel},
  booktitle={Proceedings of the 7th BlackboxNLP Workshop: Analyzing and Interpreting Neural Networks for NLP},
  pages={278--300},
  year={2024},
  month={November}
}

@article{dunefsky2024transcoders,
  title={Transcoders find interpretable llm feature circuits},
  author={Dunefsky, Jacob and Chlenski, Philippe and Nanda, Neel},
  journal={Advances in Neural Information Processing Systems},
  volume={37},
  pages={24375--24410},
  year={2024}
}

@article{lindsey2025biology,
  title={On the Biology of a Large Language Model},
  author={Lindsey, Jack and Gurnee, Wes and Ameisen, Emmanuel and Chen, Brian and Pearce, Adam and Turner, Nicholas L. and Citro, Craig and Abrahams, David and Carter, Shan and Hosmer, Basil and Marcus, Jonathan and Sklar, Michael and Templeton, Adly and Bricken, Trenton and McDougall, Callum and Cunningham, Hoagy and Henighan, Thomas and Jermyn, Adam and Jones, Andy and Persic, Andrew and Qi, Zhenyi and Thompson, T. Ben and Zimmerman, Sam and Rivoire, Kelley and Conerly, Thomas and Olah, Chris and Batson, Joshua},
  journal={Transformer Circuits Thread},
  year={2025},
  url={https://transformer-circuits.pub/2025/attribution-graphs/biology.html}
}

@article{elhage2022superposition,
   title={Toy Models of Superposition},
   author={Elhage, Nelson and Hume, Tristan and Olsson, Catherine and Schiefer, Nicholas and Henighan, Tom and Kravec, Shauna and Hatfield-Dodds, Zac and Lasenby, Robert and Drain, Dawn and Chen, Carol and Grosse, Roger and McCandlish, Sam and Kaplan, Jared and Amodei, Dario and Wattenberg, Martin and Olah, Christopher},
   year={2022},
   journal={Transformer Circuits Thread},
   url={https://transformer-circuits.pub/2022/toy_model/index.html}
}

@article{bricken2023monosemantic,
  title   = {Towards Monosemanticity: Decomposing Language Models With Dictionary Learning},
  author  = {Bricken, Trenton and Templeton, Adly and Batson, Joshua and Chen, Brian and Jermyn, Adam and Conerly, Tom and Turner, Nicholas L. and Anil, Cem and Denison, Carson and Askell, Amanda and Lasenby, Robert and Wu, Yifan and Kravec, Shauna and Schiefer, Nicholas and Maxwell, Tim and Joseph, Nicholas and Tamkin, Alex and Nguyen, Karina and McLean, Brayden and Burke, Josiah E. and Hume, Tristan and Carter, Shan and Henighan, Tom and Olah, Chris},
  year    = {2023},
  month   = {October},
  url     = {https://transformer-circuits.pub/2023/monosemantic-features},
  publisher = {Anthropic / Transformer Circuits Thread},
  journal = {Transformer Circuits Thread}
}

@article{rajamanoharan2024improving,
  title={Improving dictionary learning with gated sparse autoencoders},
  author={Rajamanoharan, Senthooran and Conmy, Arthur and Smith, Lewis and Lieberum, Tom and Varma, Vikrant and Kram{\'a}r, J{\'a}nos and Shah, Rohin and Nanda, Neel},
  journal={arXiv preprint arXiv:2404.16014},
  year={2024}
}

@article{rajamanoharan2024jumping,
  title={Jumping ahead: Improving reconstruction fidelity with jumprelu sparse autoencoders},
  author={Rajamanoharan, Senthooran and Lieberum, Tom and Sonnerat, Nicolas and Conmy, Arthur and Varma, Vikrant and Kram{\'a}r, J{\'a}nos and Nanda, Neel},
  journal={arXiv preprint arXiv:2407.14435},
  year={2024}
}

@article{ayonrinde2024adaptive,
  title={Adaptive sparse allocation with mutual choice \& feature choice sparse autoencoders},
  author={Ayonrinde, Kola},
  journal={arXiv preprint arXiv:2411.02124},
  year={2024}
}

@article{paulo2025transcoders,
  title={Transcoders beat sparse autoencoders for interpretability},
  author={Paulo, Gon{\c{c}}alo and Shabalin, Stepan and Belrose, Nora},
  journal={arXiv preprint arXiv:2501.18823},
  year={2025}
}

@article{lindsey2025landscape,
   author={Lindsey, Jack and Ameisen, Emmanuel and Nanda, Neel and Shabalin, Stepan and Piotrowski, Mateusz and McGrath, Tom and Hanna, Michael and Lewis, Owen and Tigges, Curt and Merullo, Jack and Watts, Connor and Paulo, Gonçalo and Batson, Joshua and Gorton, Liv and Simon, Elana and Loeffler, Max and McDougall, Callum and Lin, Johnny},
   title={The Circuits Research Landscape: Results and Perspectives},
   journal={Neuronpedia},
   year={2025},
   url={https://neuronpedia.org/graph/info}
}

@article{draye2026cltforge,
  author={Draye, Florent and Palit, Vedant and Harrasse, Abir and Wu, Tung-Yu and Liu, Jiarui and Pandey, Punya Syon and Wu, Roderick and Hsu, Chih-Hao and Zhang, Terry Jingchen and Jin, Zhijing and Sch{\"o}lkopf, Bernhard},
  title={{CLT-Forge}: A Scalable Library for Cross-Layer Transcoders and Attribution Graphs},
  journal={arXiv preprint arXiv:2603.21014},
  year={2026}
}

@misc{lange2026clt,
  author={Lange, Georg and {RGRGRG} and Dearstyne, Kat and Maher, Kamal},
  title={Cross-Layer Transcoders are Incentivized to Learn Unfaithful Circuits},
  howpublished={LessWrong},
  year={2026},
  url={https://www.lesswrong.com/posts/6CS2NDmoLCFcEJMor/cross-layer-transcoders-are-incentivized-to-learn-unfaithful}
}

@techreport{mcdougall2025gemmascope2,
  author={McDougall, Callum and Conmy, Arthur and Kram{\'a}r, J{\'a}nos and Lieberum, Tom and Rajamanoharan, Senthooran and Nanda, Neel},
  title={{Gemma Scope 2}: Technical Paper},
  institution={Google DeepMind},
  year={2025},
  url={https://storage.googleapis.com/deepmind-media/DeepMind.com/Blog/gemma-scope-2-helping-the-ai-safety-community-deepen-understanding-of-complex-language-model-behavior/Gemma_Scope_2_Technical_Paper.pdf}
}
\bibliographystyle{iclr2027_conference}

\clearpage

\beginappendix

\section{Auxiliary Loss}
\label{app:Aux_loss}

The auxiliary loss follows \citet{gao2025scaling}: dead latents are asked to reconstruct the error left by the live latents, which gives them a gradient signal and brings them back into use. We adapt it to cross-layer selection by applying the same pooled top-$K$ to dead latents only.

Let $\mathcal{D}$ be the set of index pairs $(i, l)$ whose latent has not fired in the last $w$ optimizer steps. Using the scores $s^l_i$ of \Cref{sec:cross-layer-topk}, we restrict selection to dead latents,
\[
\tilde{s}^l_i =
\begin{cases}
s^l_i, & (i, l) \in \mathcal{D}, \\
-\infty, & \text{otherwise,}
\end{cases}
\]
and let $\mathcal{S}'$ be the set of the $K$ index pairs with the largest $\tilde{s}^l_i$. The auxiliary activations keep the pre-activations at these positions,
\[
b^l_i =
\begin{cases}
z^l_i, & (i, l) \in \mathcal{S}', \\
0, & \text{otherwise,}
\end{cases}
\]
and are decoded with the same decoder weights as the main reconstruction,
\[
\hat{\mathbf{e}}^l = \sum_{l' \le l} \mathbf{W}_{\text{dec}}^{l' \to l} \, \mathbf{b}^{l'} .
\]
With $\mathbf{e}^l = \mathbf{y}^l - \hat{\mathbf{y}}^l$ the reconstruction error of layer $l$, the auxiliary loss is
\[
\mathcal{L}_{\mathrm{aux}} = \sum_{l=1}^{L} \big\lVert \mathbf{e}^l - \hat{\mathbf{e}}^l \big\rVert_2^2 .
\]

\section{Training Settings}
\label{app:training_setting}

\subsection{Activation Buffer}

Activations are generated on the fly. For each training sequence we run Qwen3-8B-Base and record, for every token, the MLP input and output at all $L$ layers; we call these $2L$ vectors an activation group. We sample $512$ groups uniformly at random from each sequence (all groups if the sequence is shorter), discard the rest, and add them to a buffer of $163{,}840 = 10 \times 2^{14}$ groups. Each micro-batch draws $8192$ groups uniformly at random from the buffer without replacement, and the buffer is refilled to capacity before the next micro-batch is drawn. Sampling a small subset of tokens per sequence and mixing them in a large buffer ensures that every micro-batch contains tokens from many different sequences.

\subsection{Optimization}

All models, across latent sizes and activation functions, are trained on $3.2$B tokens with identical optimization hyperparameters; we chose this budget because the reconstruction loss had plateaued by then. We use AdamW~\citep{loshchilov2019decoupledweightdecayregularization} with $(\beta_1,\beta_2) = (0.9, 0.999)$ and a weight decay of $0.01$. Gradients are accumulated over $4$ micro-batches of $8192$ tokens, giving an effective batch of $32{,}768$ tokens and about $9.8 \times 10^4$ optimizer steps in total. All schedules below are indexed by the micro-batch count $t$.

The learning rate follows a square-root warmup and then a linear decay. Over the first $100$ micro-batches ($0.8$M tokens), it rises from $\eta_0 = 0.1\eta_{\max}$ to $\eta_{\max} = 8 \times 10^{-5}$ as $\eta(t) = \eta_0 + (\eta_{\max} - \eta_0)\sqrt{t/100}$; it then decays linearly back to $\eta_0$ over the remainder of training.

\subsection{Initialization}

Our initialization setup generally follows~\citet{conerly2025dictionarylearning} and~\citet{ameisen2025circuit}. We separately normalize the activations of the input and output of each MLP layer to $\sqrt{n}$, i.e., $\mathbb{E}\left[\|X\|\right] = \sqrt{n}$, where $X$ denotes all input or output activations of the MLP in a given layer.

After normalization, we filter out groups of activations where the norm of the input or output at any layer exceeds 20 times the batch-wise mean, as such outliers would destabilize training. We observe that this filtering strategy almost exclusively acts on groups corresponding to the first token of each training sample.

We initialized every entry of each $\mathbf{W}_{\text{enc}}^l$ from $\mathcal{U}(-\frac{1}{\sqrt{n}}, \frac{1}{\sqrt{n}})$ and every entry of each $\mathbf{W}_{\text{dec}}^{l' \to l}$ from $\mathcal{U}(-\frac{1}{\sqrt{nL}}, \frac{1}{\sqrt{nL}})$.

For \CLTK, we set the per-token budget to $K = 10L = 360$, where $L = 36$ for Qwen3-8B-Base. We applied a warmup schedule to $K$, linearly decaying it from $3600$ to $360$ over the first $10^5$ micro-batches ($0.82$B tokens, about a quarter of training), after which it is held constant. The same schedule on the total budget is applied to all TopK variants. The auxiliary selection (\Cref{app:Aux_loss}) always uses the same $K$ as \CLTK.

We set the coefficient of the auxiliary loss to $\alpha=1/32$ throughout the entire training process. A latent counts as dead for the auxiliary loss if it has not fired in the last $w = 250$ optimizer steps, the same window used to report dead latents in \Cref{tab:reconstruction}.

\subsection{Baseline Hyperparameters}

\Cref{tab:baseline_hparams} lists the settings specific to each baseline; all other settings are shared with \CLTK. JumpReLU uses the sparsity penalty $\lambda \frac{2}{L_0^{\text{target}}} \big( \lVert \mathbf{a} \rVert_0 - L_0^{\text{target}} \big)^2$ described in \Cref{sec:reconstruction}.

\begin{table}[h]
\centering
\caption{Hyperparameters specific to each baseline.}
\label{tab:baseline_hparams}
\begin{tabular}{ll}
\toprule
\multicolumn{2}{l}{\textbf{JumpReLU}} \\
Target $L_0^{\text{target}}$ & $360$ \\
Penalty coefficient $\lambda$ & $0 \to 5$, linear over the first $10^5$ micro-batches \\
Straight-through estimator bandwidth & $1.0$ \\
Threshold initialization & $e^{-3.5}$ \\
Auxiliary loss coefficient~\citep{conerly2025dictionarylearning} & $3 \times 10^{-6}$ \\
\midrule
\multicolumn{2}{l}{\textbf{\CLBTK}} \\
Batches used for threshold estimation & $1{,}000$ micro-batches \\
\bottomrule
\end{tabular}
\end{table}

\section{Graph Score Definitions}
\label{app:graph_scores}

This appendix gives the definitions of the replacement score and completeness score used in \Cref{sec:interpretability}, following \citet{ameisen2025circuit}.

\paragraph{Attribution graph.}
For a prompt and a target logit $t$, the attribution graph has node set $V = E \cup F \cup R \cup \{t\}$: the embedding nodes $E$ (one per prompt token), the feature nodes $F$ (one per active CLT feature at each token position), the error nodes $R$ (one per layer and token position, absorbing the part of the MLP output not reconstructed by the CLT), and the target logit. Let $A \in \mathbb{R}^{|V| \times |V|}$ be the adjacency matrix, where $A_{v,u}$ is the direct linear effect of node $u$ on node $v$ in the replacement model. Embedding and error nodes have no incoming edges; the logit has no outgoing edges. We take $t$ to be the logit of the top-1 predicted token, which by our prompt filter (\Cref{sec:interpretability}) is the ground-truth next token.

\paragraph{Normalized influence.}
Let $\hat{A}$ be the unsigned, row-normalized adjacency matrix,
\[
\hat{A}_{v,u} = \frac{|A_{v,u}|}{\sum_{u'} |A_{v,u'}|},
\]
so that the incoming edge weights of every node with inputs sum to one. The indirect influence matrix
\[
B = (I - \hat{A})^{-1} - I = \hat{A} + \hat{A}^2 + \hat{A}^3 + \cdots
\]
sums the weights of all paths from $u$ to $v$, so $B_{v,u}$ is the total influence of $u$ on $v$ through paths of any length.

\paragraph{Replacement score.}
Because every path ends at a source node, which is either an embedding or an error node, the influence on the logit splits as $\sum_{e \in E} B_{t,e} + \sum_{r \in R} B_{t,r} = 1$. The replacement score is the fraction that originates from embeddings,
\[
\mathrm{Replacement} = \sum_{e \in E} B_{t,e},
\]
i.e.\ the fraction of end-to-end path weight from the prompt to the logit that passes only through feature nodes.

\paragraph{Completeness score.}
For each node $v$ with incoming edges (the feature nodes and the logit), let
\[
f_v = \sum_{u \in E \cup F} \hat{A}_{v,u}
\]
be the fraction of its input that comes from non-error nodes. The completeness score averages $f_v$ over nodes, weighting each node by its influence on the target logit,
\[
\mathrm{Completeness} = \frac{\sum_{v} B_{t,v}\, f_v}{\sum_{v} B_{t,v}},
\]
with the logit itself included at weight one. Unlike the replacement score, which requires a complete error-free path, the completeness score gives partial credit to a graph whose important nodes are mostly, but not entirely, explained by features.

\section{Sparse Decode Details}
\label{app:sparse_decode}

This appendix gives the algorithm for the sparse decoding scheme of \Cref{sec:efficiency}.

\paragraph{Notation.}
Following the convention of transcoder implementation notes~\citep{lindsey2025landscape}, we refer to the two sharded axes as the feature axis $F$ and the model dimension $D$, which are the latent size $m$ and the model dimension $n$ of \Cref{CLT}. $B$ is the number of tokens in a micro-batch, $R$ the number of ranks, $L$ the number of layers, and $K$ the number of active latents per token. $F_{\mathrm{loc}} = F / R$ and $D_{\mathrm{loc}} = D / R$ denote the shard held by one rank. Layers are indexed from $0$ to $L-1$, and $\mathrm{nnz}_i$ is the number of active latents that layer $i$ contributes to a micro-batch.

\paragraph{Algorithm.}
\Cref{alg:sparse_decode} gives the pseudocode of one training step. The encoder is applied to the local feature shard, and each rank proposes its $K$ highest-scoring latents per token as candidates. Since the global top-$K$ is contained in the union of the local top-$K$ sets, gathering the candidate scores ($RK$ per token) suffices to find each token's global cutoff exactly; each rank then keeps its latents above the cutoff. The active latents of all ranks are gathered in COO format by an all-reduce over four tensors of length $\mathrm{nnz}$ (values, token index, feature index, layer index), so that every rank holds the full sparse activation. Each rank then decodes into its own slice of the model dimension, and the per-token loss is all-reduced at the end. To avoid performance cliffs if sparsity collapses, each layer falls back to \texttt{dense} when $\mathrm{nnz}_i > \rho B F$, where $\rho$ is a user-defined maximum density for the sparse kernels.

\begin{algorithm}[t]
\caption{Sparse decode for one CLT training step on $R$ ranks}
\label{alg:sparse_decode}
\begin{algorithmic}[1]
\Require MLP inputs $X$, encoder shard $W_{\mathrm{enc}}$ of shape $(L, F_{\mathrm{loc}}, D)$,
decoder shards $W_{\mathrm{dec}}$ with $W_{\mathrm{dec}}[i]$ of shape $(F, L-i, D_{\mathrm{loc}})$,
kernel \texttt{mode}, one of \texttt{dense}, \texttt{gather} or \texttt{embedding\_bag}
\Statex
\Procedure{TrainStep}{$X$}
    \State $\mathrm{pre} \gets$ \textproc{Einsum}(\texttt{bld,lfd->blf}, $X$, $W_{\mathrm{enc}}$)
        \Comment{$(B, L, F_{\mathrm{loc}})$}
    \State $\mathrm{cand} \gets$ \textproc{TopK}(\textproc{Score}($\mathrm{pre}$), $K$)
        \Comment{local candidates, $(B, K)$}
    \State $\tau \gets$ \textproc{KthLargest}(\textproc{AllGather}($\mathrm{cand}$), $K$)
        \Comment{per-token global cutoff, $(B)$}
    \State $\mathrm{acts} \gets \mathrm{pre} \odot \mathbb{1}[\textproc{Score}(\mathrm{pre}) \ge \tau]$
        \Comment{exact global top-$K$}
    \State $\mathrm{sparse} \gets$ \textproc{AllReduceSum}(\textproc{Coo}($\mathrm{acts}$, \texttt{feat\_offset} $= \mathrm{rank} \cdot F_{\mathrm{loc}}$))
        \Comment{four tensors of length $\mathrm{nnz}$}
    \State \Return \textproc{SparseDecode}($\mathrm{sparse}$, $W_{\mathrm{dec}}$, \texttt{mode})
        \Comment{$(B, L, D_{\mathrm{loc}})$}
\EndProcedure
\Statex
\Procedure{SparseDecode}{$\mathrm{sparse}$, $W_{\mathrm{dec}}$, \texttt{mode}}
    \State $\mathrm{vals}, \mathrm{rows}, \mathrm{cols}, \mathrm{slices} \gets$ \textproc{SortByLayer}($\mathrm{sparse}$)
        \Comment{$\mathrm{slices}[i]$ is the range of layer $i$}
    \State $\mathrm{recon} \gets \mathbf{0}^{B \times L D_{\mathrm{loc}}}$
    \For{$i = 0$ \textbf{to} $L-1$}
        \State $\mathrm{start}, \mathrm{end} \gets \mathrm{slices}[i]$
        \State $\mathrm{row}, \mathrm{col}, \mathrm{val} \gets \mathrm{rows}[\mathrm{start}:\mathrm{end}], \mathrm{cols}[\mathrm{start}:\mathrm{end}], \mathrm{vals}[\mathrm{start}:\mathrm{end}]$
        \State $V \gets$ \textproc{View}($W_{\mathrm{dec}}[i]$, $(F, (L-i) D_{\mathrm{loc}})$)
        \State $\mathrm{out} \gets \mathrm{recon}[:, i D_{\mathrm{loc}} :]$
            \Comment{view of layers $i$ to $L-1$}
        \If{\texttt{mode} $=$ \texttt{dense}}
            \State $\mathrm{out} \gets \mathrm{out} + {}$\textproc{Scatter}($\mathrm{row}$, $\mathrm{col}$, $\mathrm{val}$, $B$, $F$)\,$V$
                \Comment{materialize $A_i$}
        \ElsIf{\texttt{mode} $=$ \texttt{gather}}
            \State \textproc{IndexAdd}($\mathrm{out}$, $\mathrm{row}$, $V[\mathrm{col}] \odot \mathrm{val}$)
                \Comment{$(\mathrm{nnz}_i, (L-i) D_{\mathrm{loc}})$ temporary}
        \Else
            \State $\mathrm{out} \gets \mathrm{out} + {}$\textproc{EmbeddingBag}($\mathrm{col}$, $V$, \textproc{Offsets}($\mathrm{row}$), $\mathrm{val}$, \texttt{sum})
                \Comment{fused}
        \EndIf
    \EndFor
    \State \Return \textproc{View}($\mathrm{recon}$, $(B, L, D_{\mathrm{loc}})$)
\EndProcedure
\end{algorithmic}
\end{algorithm}

\paragraph{FLOP counts.}
We charge $2MNP$ for an $(M\times N)(N\times P)$ matmul, plus one such product in the
backward pass for each input that requires a gradient.
The encoder einsum is evaluated \emph{once} per step---the auxiliary pre-activations
$z^{\mathrm{aux}} = z - a$ are elementwise---and its input is a received activation
buffer that carries no gradient, so only $\nabla W_{\mathrm{enc}}$ is charged:
\begin{equation}
  C_{\mathrm{enc}}
  = \underbrace{2BLF_{\mathrm{loc}}D}_{\text{forward}}
  + \underbrace{2BLF_{\mathrm{loc}}D}_{\nabla W_{\mathrm{enc}}}
  = 4BLF_{\mathrm{loc}}D ,
  \qquad F_{\mathrm{loc}} = F/W, \quad D_{\mathrm{loc}} = D/W .
\end{equation}
The decoder is evaluated \emph{twice} per step, once for the main and once for the
auxiliary reconstruction, and both its input and its weights carry gradients. Together
these contribute a factor
$\underbrace{2}_{\text{main}+\text{aux}}\times\underbrace{3}_{\text{fwd}+\text{bwd}} = 6$
over a single forward pass. A feature at source layer $i$ writes to the $L-i$ target
layers $i,\dots,L-1$, so with the \texttt{dense} kernel
\begin{equation}
\begin{aligned}
  C_{\mathrm{dec}}^{\mathrm{dense}}
  &= 6\sum_{i=0}^{L-1} 2BF(L-i)D_{\mathrm{loc}}
   = 12\,BFD_{\mathrm{loc}}\,\frac{L(L+1)}{2}
   = \frac{6BFDL(L+1)}{W}, \\[2pt]
  C_{\mathrm{step}}^{\mathrm{dense}}
  &= C_{\mathrm{enc}} + C_{\mathrm{dec}}^{\mathrm{dense}}
   = \frac{2BLFD}{W}\,(3L+5),
\end{aligned}
\end{equation}
the same decoder FLOP count as when sharding along $F$. Sharding along $D$ does,
however, narrow the reconstruction accumulator from $D$ to $D_{\mathrm{loc}}$, cutting
its $BDL(L+1)$ words of traffic per decode by a factor of $W$, and it removes the
reconstruction all-reduce, saving $2BLD$ words of communication per step. The
\texttt{dense} kernel pays part of this back by re-densifying the gathered activations
through a $2BFL$-word scratch buffer, so the \emph{total} per-rank decoder traffic
falls by $2.2\times$ at our configuration rather than by $W$.
With the \texttt{gather} and \texttt{embedding\_bag} kernels only the active rows are
read. Writing $\mathrm{nnz}_i$ for the number of selected features at source layer $i$
and $K = Lk$ for the per-token budget, the same prefactor $6$ gives
\begin{equation}
\begin{aligned}
  C_{\mathrm{dec}}^{\mathrm{sparse}}
  &= 6\sum_{i=0}^{L-1} 2\,\mathrm{nnz}_i\,(L-i)\,D_{\mathrm{loc}}
   \;\approx\; 12\,BkD_{\mathrm{loc}}\,\frac{L(L+1)}{2}
   = \frac{6BKD(L+1)}{W}, \\[2pt]
  C_{\mathrm{step}}^{\mathrm{sparse}}
  &= \frac{2BD}{W}\bigl(2LF + 3K(L+1)\bigr),
\end{aligned}
\end{equation}
replacing the factor $FL$ in the decoder term by $K$.

\end{document}